\documentclass[conference]{IEEEtran}
\IEEEoverridecommandlockouts
\usepackage{amsmath,amsfonts}
\usepackage{algorithmic}
\usepackage{algorithm}
\usepackage{array}
\usepackage{textcomp}
\usepackage{stfloats}
\usepackage{url}
\usepackage{verbatim}
\usepackage{graphicx}
\usepackage{cite}
\usepackage{xcolor}
\usepackage{booktabs}
\newcommand{\ra}[1]{\renewcommand{\arraystretch}{#1}}
\usepackage{csvsimple}
\usepackage{hyperref}
\usepackage[capitalise]{cleveref}
\usepackage{bm}
\usepackage{xspace}
\usepackage{subcaption}

\newcommand{\M}[1]{{\bm #1}} %

\newcommand{\SEthree}{\ensuremath{\mathrm{SE}(3)}\xspace}

\newcommand{\SOthree}{\ensuremath{\mathrm{SO}(3)}\xspace}

\newcommand{\numberthis}{\addtocounter{equation}{1}\tag{\theequation}}

\def\editing{1}
\if \editing1

\fi

\newcommand{\methodname}{PEOPLEx\xspace}

\begin{document}

\title{\methodname : PEdestrian Opportunistic Positioning\\ LEveraging IMU, UWB, BLE and WiFi}

\author{Pierre-Yves Lajoie, Bobak Hamed Baghi, Sachini Herath, Francois Hogan, Xue Liu, Gregory Dudek
\\\IEEEauthorblockA{Samsung Electronics, Canada \\ \{p.lajoie,s.herath\}@partner.samsung.com, \{bobak.h, f.hogan, steve.liu, greg.dudek\}@samsung.com}%
}

\maketitle

\begin{figure*}[hb]
    \vspace{-5mm}
    \centering
    \includegraphics[width=\textwidth]{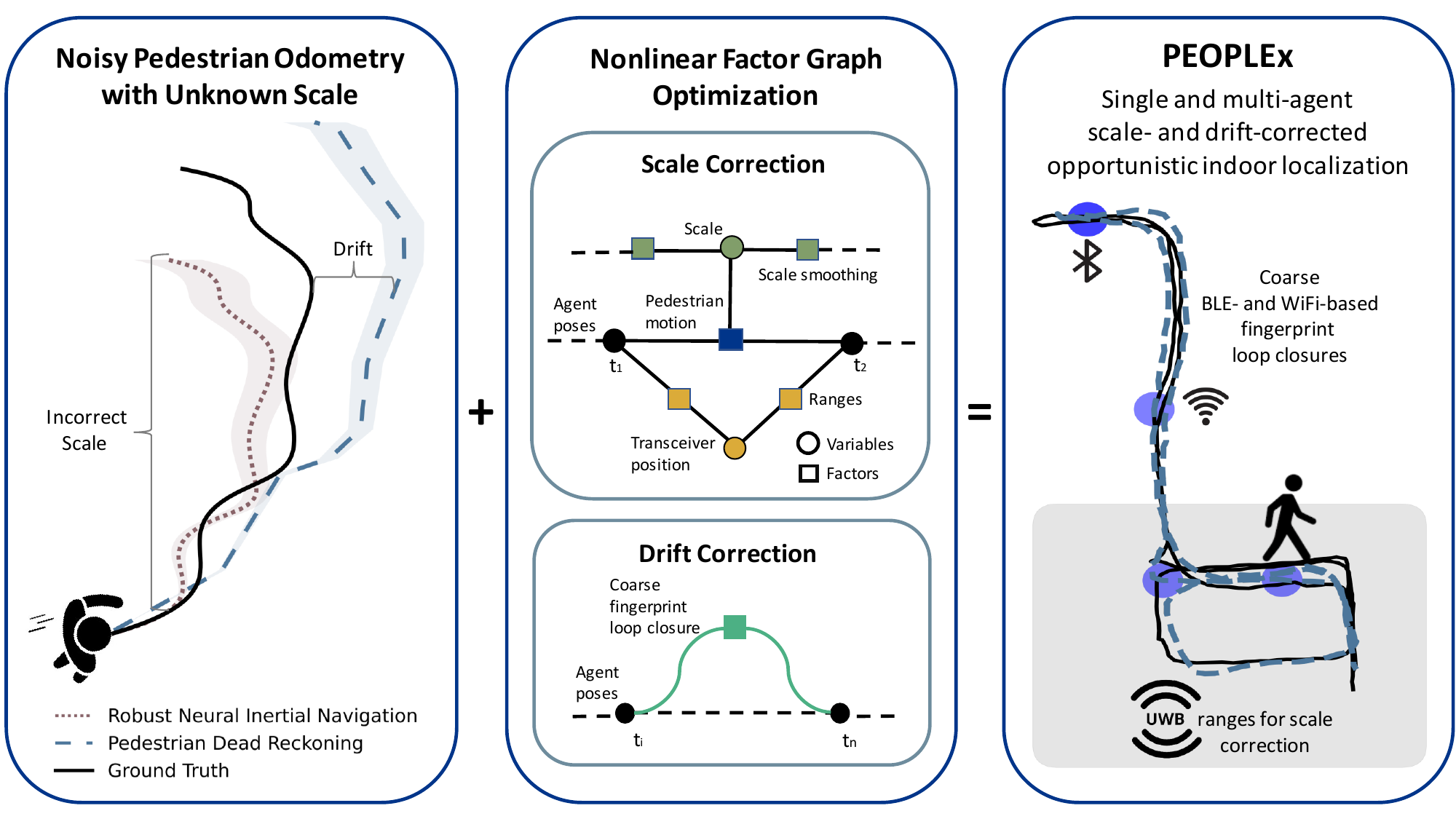}
    \caption{Illustration of the \methodname framework fusing sensor data from multiple sources for opportunistic pedestrian localization.}
    \label{fig:top}
\end{figure*}

\begin{abstract}
This paper advances the field of pedestrian localization by introducing a unifying framework for opportunistic positioning based on nonlinear factor graph optimization. 
While many existing approaches assume constant availability of one or multiple sensing signals,
our methodology employs IMU-based pedestrian inertial navigation as the backbone for sensor fusion, opportunistically integrating Ultra-Wideband (UWB), Bluetooth Low Energy (BLE), and WiFi signals when they are available in the environment.
The proposed \methodname framework is designed to incorporate sensing data as it becomes available, operating without any prior knowledge about the environment (e.g. anchor locations, radio frequency maps, etc.).
Our contributions are twofold: 1) we introduce an opportunistic multi-sensor and real-time pedestrian positioning framework fusing the available sensor measurements; 2) we develop novel factors for adaptive scaling and coarse loop closures, significantly improving the precision of indoor positioning. Experimental validation confirms that our approach achieves accurate localization estimates in real indoor scenarios using commercial smartphones.

\end{abstract}

\vspace{5mm}
\section{Introduction}
\IEEEPARstart{W}{ith} an ever-growing number of interconnected devices and systems, effective and accurate indoor positioning has become increasingly critical. The real-time location of the user is an valuable data point that heavily impacts the \textit{state} of the environment in both household and industrial settings, and can influence the interpretation of other data streams as well as any potential action to be undertaken by the IoT system. 
Indoor Positioning Systems (IPS) hold the potential to transform the operations and experiences of pedestrians, underpinning applications such as navigation assistance, location-based services, safety enhancement, and augmented reality experiences.

Historically, building reliable indoor positioning solutions often necessitates prior knowledge of the environment, such as the precise locations of Radio Frequency (RF) anchors~\cite{Pan2022}, floor plans~\cite{Herath2021}, and more~\cite{zafariSurveyIndoorLocalization2019}. However, acquiring and utilizing such comprehensive information is not always feasible, nor desirable, due to issues related to availability, operational cost, and privacy concerns.
Recognizing these challenges, our research focuses on smartphone-based localization strategies that use exclusively the sensing modalities available in the environment, are allowed by the users, and do not require prior knowledge about the environment.
Smartphones, which are ubiquitous, offer a rich set of built-in sensors and radios (e.g., IMUs, Wi-Fi, Bluetooth Low Energy (BLE), Ultra-Wideband (UWB)) that can be utilized for localization. Unlike cameras, also available in smartphones, those sensors do not require active supervision and raise less privacy concerns.

The method proposed in this paper, \methodname, leverages pedestrian inertial navigation to fuse multiple sensor modalities, capitalizing on whatever sensor data is readily available at any given time.
To achieve this, we introduce a methodology that uses nonlinear factor graph optimization as a unifying framework to integrate information from the diverse sensor modalities available on smartphones. At the center of this approach is the inertial motion estimation, which produces noisy and up-to-scale pedestrian trajectories, effectively acting as the `glue' that ties together the input from other sensors like Wi-Fi, BLE, and UWB.
Utilizing these trajectories, our nonlinear factor graph approach enables simultaneous optimization of both UWB anchor positions and user locations, eliminating the necessity for predefined initial knowledge of the environment.
\methodname integrates two distinct forms of IMU-based pedestrian inertial navigation, both agnostic to sensor-placement on the body, thereby offering a comprehensive analysis of the approach's effectiveness and adaptability.
In essence, our novel
formulation, illustrated in~\cref{fig:top}, allows us to simultaneously estimate key motion parameters, such as step length or scaling, localize the user, and construct a coarse map of RF sources. 

In summary, we present the following contributions:
\begin{itemize}
        \item An opportunistic framework that leverages pedestrian inertial navigation to fuse sensor measurements, without requiring any initial knowledge about the environment such as anchor locations or radio frequency maps;
        \item Novel factor formulations for adaptive scaling and coarse loop closing to robustly integrate data from RF sensors, enhancing the precision of indoor positioning;
\end{itemize}
We validate our contributions in extensive experiments with smartphone IMUs and two exterosceptive sensing mechanism: WiFi and BLE fingerprinting, and UWB ranging.

\section{Background and Related Work}

\subsection{Pedestrian Inertial Navigation}
Pedestrian inertial navigation is a central component of indoor positioning research due to its independence from external infrastructures and its relatively low computational requirements. It leverages the inertial measurement units (IMU), available in many consumer devices to track pedestrian motion, estimating position based on parameters such as step length, velocity and heading direction~\cite{harleSurveyIndoorInertial2013}. Recent studies have used learning-based approaches towards refining the accuracy and reliability, exploiting the regularity of pedestrian motion to infer the overall user motion from the high frequency and high noise accelerometer and gyroscope data~\cite{herathRoNINRobustNeural2020}.

Despite these advancements, those approaches still face notable challenges, most significantly the issue of accumulated error over time. This drift error arises because the estimation process incrementally incorporates motion estimates from one step to another without correcting the noise from previous steps. Therefore, small inaccuracies compound over time leading to significant positioning errors~\cite{harleSurveyIndoorInertial2013}. 

\subsection{RF-based Indoor Positioning Systems}

Radio Frequency (RF)-based systems have become a popular solution for indoor positioning due to their capability to provide relatively accurate positioning based on existing infrastructure in indoor environments~\cite{zafariSurveyIndoorLocalization2019}. There exists a variety of RF technologies utilized for indoor positioning, including Wi-Fi, Bluetooth Low Energy (BLE), and Ultra-Wideband (UWB), each with their unique strengths and limitations.

Wi-Fi based positioning systems are among the most common and cost effective due to the widespread availability of Wi-Fi infrastructure (e.g. routers, consumer devices, etc.). They typically employ techniques such as Received Signal Strength Indicator (RSSI) fingerprinting to estimate devices positions. However, unless extensive and costly acquisition of fingerprint maps is undertaken, Wi-Fi-based systems tend to suffer from signal interference, often leading to suboptimal localization accuracy~\cite{shangOverviewWiFiFingerprintingbased2022}.
With the increasing number of connected devices, BLE fingerprint-based systems have emerged as a suitable alternative to Wi-Fi. BLE beacons can provide positioning data with a small energy footprint, making them particularly suitable for battery-powered devices. However, they can be affected by signal instability and require a dense beacon deployment for optimal performance~\cite{faragherLocationFingerprintingBluetooth2015}.

Ultra-Wideband (UWB) positioning systems stand out due to their high precision and resilience to multipath effects. 
UWB systems often utilize Two-Way Ranging (TWR), since it does not necessitate clock synchronization between the devices involved, thus reducing system complexity while making it easier to deploy~\cite{baruaPerformanceStudyTWR2018}.
In conventional indoor positioning using (single antenna) UWB, at least three to four anchors with known locations are typically needed for 2D or 3D estimation. The system's precision is also sensitive to the placement of these sensors, deteriorating when the anchors are sub-optimally located~\cite{zhaoFindingRightPlace2022}.
Unlike most UWB-based techniques, our approach does not assume any prior knowledge of UWB anchors placement, and can improve positioning estimates using a single UWB transceiver.

\subsection{Sensor Fusion Approaches}
The importance of sensor fusion, the process of combining data from multiple sensory sources, has gained substantial recognition in the domain of indoor positioning. By coalescing information from various sensors, these methods strive to harness the advantages and counterbalance the limitations of individual sensors, thereby enhancing the accuracy and robustness of positioning systems~\cite{Poulose2019}.

Numerous methodologies have been employed to implement sensor fusion. One prevalent approach is the integration of data from inertial measurement units (IMUs) with RF signals such as UWB. Conventionally, filtering techniques like Kalman filters and Particle filters have been the standard methods for merging the data from these diverse sources. While these techniques have demonstrated improvement in positioning accuracy compared to single sensor-based systems~\cite{zhongIntegrationUWBIMU2018}, they exhibit certain limitations. These methods often necessitate intricate sensor and motion modeling and typically lack the capability to reassess and refine past estimates. 

Smoothing-based formulations based on factor graph address those limitations. Factor graphs provide a flexible framework for modeling the complex dependencies between various sensor measurements and position estimates, and are able to refine past estimates as new measurements are acquired~\cite{kaessISAM2IncrementalSmoothing2012a}. 
Representing variables and constraints as nodes and edges, respectively, this approach leverages the problem's sparse structure for efficient computation, which is crucial for large-scale, long-term tasks like indoor positioning~\cite{dellaertFactorGraphs}.

\subsection{Opportunistic Approaches}

Opportunistic approaches 
have recently gained attention, providing potential solutions for scenarios where traditional localization systems may not be suitable. Unlike conventional methods, opportunistic approaches do not depend on the constant availability of specific signals or information sources but instead utilize whatever data is readily available.

While various works in the field strive to enhance inertial pedestrian navigation by fusing it with data from supplementary sensors, they often fall short of providing a comprehensive solution. For example, Tian et al.\cite{tianLowCostINSUWB2019} utilize a particle filter to combine inertial navigation and UWB range measurements from a single anchor. However, this approach demands an initial anchor position estimation and lacks the capability for continuous refinement, contrasting with our factor graph methodology. 
Liu et al.~\cite{liuCollaborativeSLAMBased2020}, although employing a factor graph approach similar to ours, make a simplifying assumption of a constant step length, thereby introducing drift and scale inaccuracies, and does not integrate range measurements.

Jao et al.~\cite{jaoUWBFootSLAMBoundingPosition2023} implement an Extended Kalman Filter to merge data from foot-mounted IMUs and UWB sensors. While foot-mounted IMUs offer enhanced tracking performance, they lack the practicality of widely available smartphone sensors.
In a similar vein, Chen et al.~\cite{chenjianfan2022} and Lu et al.~\cite{Jinjie2022} offer methodologies that necessitate prior knowledge or surveying of the environment.
Chen et al.~\cite{chenjianfan2022} fuses pedestrian inertial navigation with BLE fingerprinting and trilateration, while Lu et al.~\cite{Jinjie2022} proposes a data-driven IMU and WiFi indoor localization system.
Our approach, in contrast, does not assume any prior environmental information and provides a real-time, multi-modal sensor fusion framework capable of online parameter estimation, such as step length and anchor positions.

\section{\methodname \xspace Framework}

In this work, we leverage nonlinear factor graph optimization in conjunction with the sensing of environmental radio-frequency signals (i.e. WiFi, BLE, and UWB) to correct the scale and drift of IMU-based localization techniques.
In absence of sufficient radio-frequency signals, our technique performs inertial navigation alone which can be relied upon for short periods of time.
Thus, our approach is opportunistic in nature, enabling the use of an available RF sensing modality when possible, yet consistently delivering a solution \textemdash albeit potentially less accurate\textemdash\xspace even when it's not available.

The real power of our methodology comes from the inclusion of custom nonlinear factors tailored specifically for pedestrian indoor positioning.
These factors encode domain-specific knowledge such as unique sensor characteristics and key pedestrian parameters directly in the optimization problem.
To efficiently optimize on the Special Euclidean group \SEthree (i.e. rotation and translation of the user), nonlinear factor graph optimization linearizes the problem at the current estimate, approximating it within a tangent space. 
The linearized problem is then solved, 
 and the solution is mapped back to the original \SEthree space for variable updates~\cite{dellaertFactorGraphs}. 
The iterative process of linearization and updates is repeated until it eventually converges to a solution.
Thus, to implement our custom factors, we must specify appropriate error functions and their associated Jacobians in the \SEthree tangent space, which are described in the following subsections. 

\subsection{Adaptive Scale for Pedestrian Motion}\label{sec:adapt-factors}

In our study, we employ two types of inertial navigation techniques: step counting-based Pedestrian Dead Reckoning (PDR) and Robust Neural Inertial Navigation (RoNIN~\cite{herathRoNINRobustNeural2020}). Both techniques use accelerometers and gyroscopes to estimate relative motion. However, they are inherently prone to accumulating errors over time, i.e. drift.

On one hand, PDR counts the step of the user by detecting spikes in acceleration~\cite{Wang2019}, and estimates the heading direction of each step (i.e. yaw) by combining accelerometer and gyroscope data.
In closely related prior work~\cite{liuCollaborativeSLAMBased2020}, PDR operates under the assumption of a constant step length provided as an input parameter. This approach is not realistic, as step length can vary significantly between individuals and can also fluctuate over time even within a single trajectory. 
Solutions for step length estimation, such as foot-mounted IMUs~\cite{jaoUWBFootSLAMBoundingPosition2023}, are often not practical as they require specialized hardware.
Therefore, in our framework, we extend traditional Pedestrian Dead Reckoning (PDR) by incorporating a custom nonlinear factor to jointly estimate the step length and the agent's pose during the localization process. 

On the other hand, RoNIN~\cite{herathRoNINRobustNeural2020} integrates IMU data overtime using a neural network architecture and computes a stable and accurate relative velocity. Yet, the scale of the velocity estimates is intrinsically tied to the data used for training the network. Thus if the user is smaller, taller or has a different gait from the original data collectors, the resulting RoNIN trajectory estimates need to be scaled by a constant parameter analog to the step length in PDR.

\subsubsection{Step Counting-based Pedestrian Dead Reckoning}

We define the agent's pose before and after a step as \( \M{T}_0 \) and \( \M{T}_1 \) respectively. Both poses belong to \( \SEthree \). The measured relative rotation matrix \( \bar{\M{R}} \) from the IMU belongs to \( \SOthree \) and, finally, \( s \) and \( \boldsymbol{u}  \) are the scaling variable corresponding to the step length and the unit vector corresponding to the walking direction, respectively.

\vspace{-5mm}
\begin{align*}
\M{T}_0, \M{T}_1& \in \SEthree, \;\;
\bar{\M{R}}_{0,1} \in \SOthree, \;\;
s \in {\mathbb R} \\
\bar{\boldsymbol{t}}_{0,1} &= s \cdot \boldsymbol{u}  \numberthis
\label{eq:pdr-translation}
\end{align*}

To build the measured relative pose \( \bar{\M{T}}_{\text{step}} \) for a step, we utilize the measured rotation \( \bar{\M{R}}_{0,1} \) and the translational component \( \bar{\boldsymbol{t}}_{0,1} \) which is dependent on the scale variable \( s \).
We also define the relative pose $\M{T}_{0,1}$ between our estimates and the pedestrian motion error term \( \rho_{\text{motion}} \):

\vspace{-5mm}
\begin{align*}
\bar{\M{T}}_{\text{step}} &= \begin{bmatrix}
\bar{\M{R}}_{0,1} & \bar{\boldsymbol{t}}_{0,1}\\
0 & 1
\end{bmatrix}.\\
\M{T}_{0,1} &= \M{T}_{0}^{-1} \cdot \M{T}_{1}.\\
\rho_{\text{motion}} &= \text{Log}(\bar{\M{T}}_{\text{step}}^{-1} \cdot \M{T}_{0,1}), \numberthis
\end{align*}

where $\text{Log}$ is the logarithm map to obtain the error in tangent space~\cite{dellaertFactorGraphs}.

The Jacobian matrices are critical for the optimization process, as they tell the optimizer how a small change in each variable would affect the error. In our case, we need the Jacobians \( \M{H}^{\rho_{\text{motion}}}_{\M{T}_{0}} \), \( \M{H}^{\rho_{\text{motion}}}_{\M{T}_{1}} \), and \( \M{H}^{\rho_{\text{motion}}}_s \) of the error function \(\rho_{motion}\) with respect to the pose before the step, the pose after the step, and the scaling variable \( s \) respectively.

\vspace{-5mm}
\begin{align*}
\M{H}^{\rho_{\text{motion}}}_{\M{T}_{0}} &= - \text{Adj}(\M{T}_{0,1}^{-1}), \;\;\;\;
\M{H}^{\rho_{\text{motion}}}_{\M{T}_{1}} = \M{I},\\
\M{H}^{\rho_{\text{motion}}}_{s} &= \begin{bmatrix} 0, 0, 0, -\boldsymbol{u} \end{bmatrix}^\top.  \numberthis
\end{align*}

The Jacobian \( \M{H}^{\rho_{\text{motion}}}_{\M{T}_{0}} \) is the negative adjoint representation of the relative pose \( \M{T}_{0,1}^{-1} \). It essentially captures how a small change in the initial \SEthree pose influences the error term in tangent space. We take the negative because of the inversion operation in \( \M{T}_{0,1} \).
Since \( \M{T}_1 \) directly contributes to \( \M{T}_{0,1} \), the Jacobian \( \M{H}^{\rho_{\text{motion}}}_{\M{T}_{1}} \) is simply the identity matrix. Any change in \( \M{T}_1 \) would directly translate to the same magnitude of change in the error term.
The Jacobian  \( \M{H}^{\rho_{\text{motion}}}_{s} \) reflects how a change in the scale \( s \) would negatively impact the translational component of the error term along the walking direction $\boldsymbol{u}$.

As shown in the upper-middle section of~\cref{fig:top}, we also add a scale smoothing factor \( \rho_{\text{scale}} \) to the factor graph. This factor is used to penalize large changes between the scales $s_i$ and $s_j$ of consecutive steps. The scale smoothing factor is defined as follows:

\vspace{-5mm}
\begin{align}
    \rho_{\text{scale}} &= s_j -s_i,\;\;
    \M{H}^{\rho_{\text{scale}}}_{s_i} = -1,\;\;
    \M{H}^{\rho_{\text{scale}}}_{s_j} = 1.
\end{align}

\subsubsection{Robust Neural Inertial Navigation (RoNIN)}
In contrast to step counting-based PDR, RoNIN employs a neural network to integrate IMU data and produce robust relative velocity estimates \( \bar{\boldsymbol{v}} \) over a time interval \( \Delta_\text{time} \). RoNIN's initial velocity scale is influenced by the ground truth data used during the network training. To accommodate variations in user size, gait, and other parameters, we introduce a scaling variable \( s \) analog to the one we use for the step counting-based approach. Note that scale smoothing is also used for RoNIN.

\vspace{-5mm}
\begin{align}
    \bar{\boldsymbol{t}}_{0,1} &= s \cdot \bar{\boldsymbol{v}} \cdot \Delta_\text{time}  \label{eq:ronin-translation}
\end{align}

The relative pose \( \bar{\M{T}}_{\text{step}} \) in the RoNIN-based method is assembled using the scaled translation \( \bar{\boldsymbol{t}}_{0,1} \) in a slightly different manner compared to the PDR. In this case, the relative pose matrix represents only translation in \( x \) and \( y \) directions, \(\bar{x}_{0,1}\) and \(\bar{y}_{0,1}\), without accounting for rotation.
The relative pose error function stays the same.

\vspace{-5mm}
\begin{align*}
    \bar{\M{T}}_{\text{step}} &= \left[ \begin{array}{ *{4}{c} }
    & & & \bar{x}_{0,1} \\
    & & & \bar{y}_{0,1}  \\
    \multicolumn{3}{c}
      {\raisebox{\dimexpr\normalbaselineskip+.7\ht\strutbox-.5\height}[0pt][0pt]
        {\scalebox{2}{\textit{I}}}} & 1 \\
    0 & 0 & 0 & 0
  \end{array} \right],
  \rho_{\text{motion}} = \text{Log}(\bar{\M{T}}_{\text{step}}^{-1} \cdot \M{T}_{0,1}).\\
\M{H}^{\rho_{\text{motion}}}_{s} &= \begin{bmatrix} 0, 0, 0, -\bar{v}_x \cdot \Delta_\text{time}, -\bar{v}_y \cdot \Delta_\text{time}, 0 \end{bmatrix}^\top. \numberthis
\end{align*}

The Jacobians \( \M{H}^{\rho_{\text{motion}}}_{\M{T}_{0}} \) and \( \M{H}^{\rho_{\text{motion}}}_{\M{T}_{1}} \) are the same as in the step counting method. However, the Jacobian \( \M{H}^{\rho_{\text{motion}}}_{s} \) reflects a different sensitivity of the error term to changes in the scaling variable \( s \). It takes into account the effect on both \( x \) and \( y \) components of the velocity \( \bar{\boldsymbol{v}} \) over the time interval \( \Delta_\text{time} \).

\subsection{Drift Correction via Coarse Relocalization}\label{sec:coarse-proximity}
In our solution, we also incorporate other measurement types such as BLE (Bluetooth Low Energy) and WiFi fingerprinting, which are widely used in indoor localization~\cite{zafariSurveyIndoorLocalization2019}.
We perform fingerprinting online, which means that we collect fingerprints as we move through the environment and use them to correct the drift in our inertial navigation estimates by producing loop closures between poses (i.e., steps) sharing two highly similar fingerprints. 
Frequently used to reduce localization drift SLAM, loop closures are constraints added between non-consecutive steps that are recognized to be in the same location~\cite{loopClosure2023}.
Specifically, using consumer smartphones, we periodically perform BLE and WiFi scans to collect Received Signal Strength Indicators (RSSI), in dBm (decibel-milliwatts), from devices present in the environment. The RSSI values from a scan $i$ are stored into a fingerprint vector $f_i$ for which each entry correspond to a unique device. 
We compare fingerprint vectors from different scans (e.g. $f_i$ and $f_j$) using a cosine similarity.
If the similarity is sufficiently high, we introduce a loop closure constraint. To avoid false positive, we only consider scans with at least ten RSSI values.
 Due to limitations in fingerprinting accuracy, the discretization of steps, and the limited BLE/WiFi scan frequencies, the association of step-to-scan data is inherently imperfect. Thus, with smartphones usually constrained by low scan rates, BLE and Wi-Fi fingerprinting solutions lack reliability for precise localization. 
 While the standard approach in the literature typically employs simple proximity constraints, which can be enhanced by learning a mapping from similarity to distance as seen in~\cite{liuCollaborativeSLAMBased2020}, this approach has the drawback of being highly environment and device-specific.

To address these challenges in a way that is agnostic to the environment, we introduce a coarse loop closure method. This method defines a circular region within which loop closures are costless, while outside of this region the cost increases in function of the distance. Essentially, it encourages two estimated locations to remain close up to a certain distance. The distance threshold is conservatively set to account for the inherent inaccuracies of the fingerprinting system.

\vspace{-5mm}
\begin{align}
\M{T}_i, \M{T}_j& \in \SEthree,\;\; 
\boldsymbol{q} = \M{R}_i^\top \cdot (\boldsymbol{t}_j - \boldsymbol{t}_i).
\end{align}

Here, $\M{T}_i$ and $\M{T}_j$ are transformation matrices representing two non-consecutive poses linked by a fingerprint loop closure, and $\boldsymbol{q}$ is the relative translation vector between them in pose $\M{T}_i$ reference frame.

\vspace{-5mm}
\begin{align}
\rho_\text{loop} = \begin{cases}
\left\lVert \boldsymbol{q}  \right\rVert - r & \text{if } \left\lVert \boldsymbol{q}  \right\rVert > r, \\
0 & \text{otherwise. } 
\end{cases}
\end{align}

The loop closure cost, $\rho_\text{loop}$, is defined as follows:
if $\left\lVert \boldsymbol{q} \right\rVert$ (the Euclidean norm of $\boldsymbol{q}$) is greater than a predefined trust radius $r$, the cost is the difference between $\left\lVert \boldsymbol{q} \right\rVert$ and $r$. This encourages loop closures when poses are outside the radius, otherwise, if $\left\lVert \boldsymbol{q} \right\rVert$ is less than or equal to $r$, the cost is set to $0$, indicating that the poses are within the loop closure radius.
We define the Jacobians as follows:

\vspace{-5mm}
\begin{align*}
\boldsymbol{q}_\text{norm} &= \left[ \frac{\boldsymbol{q}_{x}}{\left\lVert \boldsymbol{q}  \right\rVert}, \frac{\boldsymbol{q}_{y}}{\left\lVert \boldsymbol{q}  \right\rVert}, \frac{\boldsymbol{q}_{z}}{\left\lVert \boldsymbol{q}  \right\rVert} \right], \\
\M{H}^{\rho_\text{loop}}_{\M{T}_{i}} &= \boldsymbol{q}_\text{norm} \cdot \begin{bmatrix} 0 & -\boldsymbol{q}_z & \boldsymbol{q}_y & -1 & 0 & 0\\
\boldsymbol{q}_z & 0 & -\boldsymbol{q}_x & 0 & -1 & 0\\
-\boldsymbol{q}_y & \boldsymbol{q}_x & 0 & 0 & 0 & -1\end{bmatrix}\!\!,\\
\M{H}^{\rho_\text{loop}}_{\M{T}_{j}} &= \left[0,0,0,  \boldsymbol{q}_\text{norm} \cdot \M{R_i}^\top \cdot \M{R}_j\right]. \numberthis
\end{align*}

where $\boldsymbol{q}_\text{norm}$ is the normalized relative translation vector, while $\M{H}^{\rho_\text{loop}}_{\M{T}_{i}}$ and $\M{H}^{\rho_\text{loop}}_{\M{T}_{j}}$ represent the sensitivity of the loop closure cost to changes in the poses $i$ and $j$.
In the cases where $\rho_\text{loop}$ is equal to $0$ (i.e., the poses are within the loop closure radius), we set the Jacobians to zero.

\begin{figure}
    \centering
    \includegraphics[width=\columnwidth]{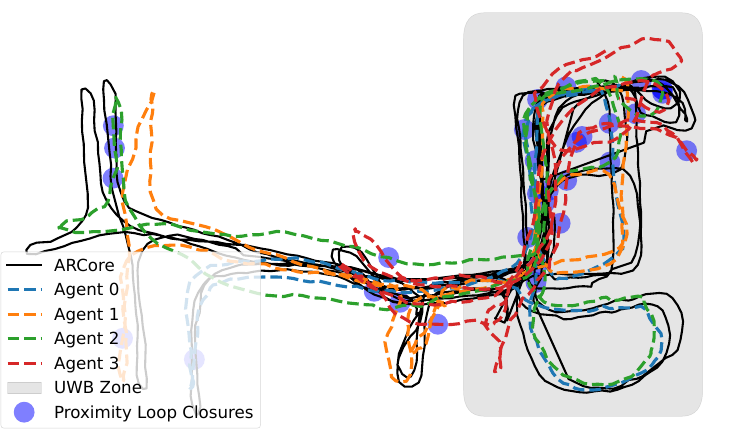}
    \caption{Illustration of the multi-agent scenario, featuring UWB ranging, BLE and WiFi coarse loop closures, and RoNIN as the pedestrian motion model. We acheive a resulting RMSE of 1.06 $\pm$ 0.53.}
    \label{fig:multi-agent}
\end{figure}

\subsection{Opportunistic Positioning Solution}

Our method results in a solution that is both robust and uniquely suited for the complex challenges of indoor pedestrian localization.
We seek to recover the optimal sequence of 3D transformations $\M{T}_0^*, \M{T}_1^*, \ldots, \M{T}_n^*$ by minimizing the weighted sum of squared residuals corresponding to the measurements. These transformations represent the poses of the pedestrian at different time steps. Specifically, we solve:

\begin{align*}
    \M{T}_0^* \dots \M{T}_n^* &\\
    \!\!\!\!\!=\! \operatorname*{argmin}_{\M{T}_0 \dots \M{T}_n} &\!\! \sum_{i \in \text{steps}}\!\!\!\! \| \rho_\text{motion}(\M{T}_i, \M{T}_{i+1}) \|^2_{\M{\Omega}_\text{motion}}\!\!\!\! +\! \| \rho_\text{scale}(s_i, s_{i+1}) \|^2_{\M{\Omega}_\text{scale}} \\
    & + \sum_{(bi, bj) \in \text{BLE loops}} \| \rho_\text{loop}(\M{T}_{bi}, \M{T}_{bj}) \|^2_{\M{\Omega}_\text{BLE}}  \\
    & + \sum_{(wi, wj) \in \text{WiFi loops}} \| \rho_\text{loop}(\M{T}_{wi}, \M{T}_{wj}) \|^2_{\M{\Omega}_\text{WiFi}} \\
    & + \sum_{(u, a, d) \in \text{UWB ranges}} \| \rho_\text{range}(\M{T}_u, \boldsymbol{t}_a; d) \|^2_{\M{W}_\text{UWB}}. \numberthis
\end{align*}

\begin{figure}
    \centering
    \includegraphics[width=\columnwidth]{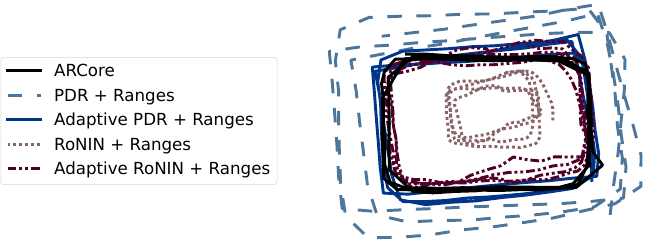}
    \caption{Scaling example: Starting from an incorrect scale (0.5 for RoNIN and 2.0 for PDR), we show that our adaptive techniques (Adaptive PDR + Ranges and Adaptive RoNIN + Ranges) are able to self-correct and obtain estimates close to the ARCore ground truth.}
    \label{fig:scaling-ex}
\end{figure}

The weight matrices $\M{\Omega}$, represented as $\M{\Omega}_\text{motion}$, $\M{\Omega}_\text{scale}$, $\M{\Omega}_\text{BLE}$, and $\M{\Omega}_\text{WiFi}$, are the information matrices (i.e., inverse of the covariance) of the corresponding measurements. 
The error term $\rho_\text{range}$ is the standard range measurement error term between a pose from the trajectory $\M{T}_u$ and the 3D position of an anchor $\boldsymbol{t}_a$, with $d$ as the measured range~\cite{dellaertFactorGraphs}. The weight matrix $\M{W}_\text{UWB}$ incorporates a Cauchy robust loss  function based on the residual value~\cite{Mlotshwa2022} to account for outliers in the collected ranges, particularly those arising from non-line-of-sight measurements. 
To this formulation is added a pose prior fixing the first pose to the origin of the world frame. 
Data from smartphones and UWB sensors are continuously collected and transmitted to a central server for real-time processing. The optimization process is performed using the Levenberg-Marquardt-based iSAM2 iterative algorithm~\cite{kaessISAM2IncrementalSmoothing2012a}

\begin{figure*}
    \begin{subfigure}[b]{0.317\textwidth}
        \centering
        \includegraphics[width=\columnwidth]{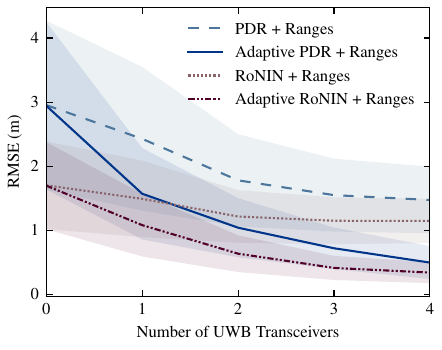}\vspace{-2mm}
        \caption{Effect of the number of UWB anchors.}
        \label{fig:nb-uwb}
    \end{subfigure}
    \begin{subfigure}[b]{0.327\textwidth}
        \centering
        \includegraphics[width=\columnwidth]{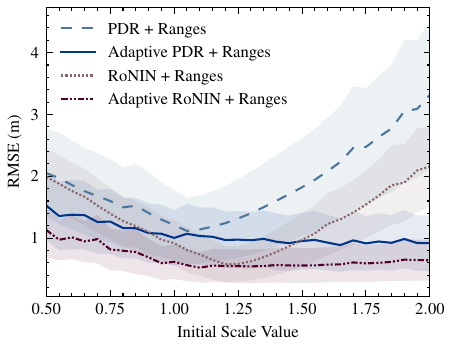}\vspace{-2mm}
        \caption{Effect of the initial scale value.}
        \label{fig:scale_uwb}
    \end{subfigure}
    \begin{subfigure}[b]{0.327\textwidth}
        \centering
        \includegraphics[width=\columnwidth]{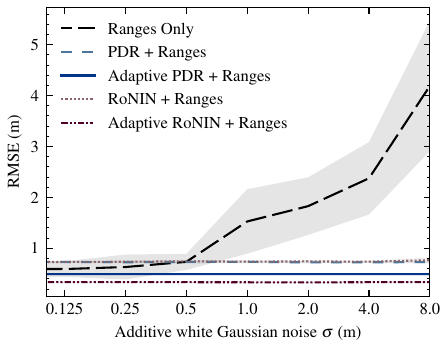}\vspace{-2mm}
        \caption{Effect of initial UWB anchors estimates}
        \label{fig:uwb-placement-noise}
    \end{subfigure}
    \caption{Comparison between fixed and adaptive scaling approaches for inertial navigation and UWB ranging solutions.}
    \vspace{-5mm}
\end{figure*}

\section{Experiments}

For the experiments, we used Samsung S22 phones carried by the pedestrian subjects. Through the Android API, we collect orientation data, accelerometer data for step counting, and both accelerometer and gyroscope data for RoNIN-based inertial navigation. Decawave DW1001 Ultra
Wideband (UWB) transceivers operating at 10~Hz were employed for ranging, with one module on each phone and four additional modules serving as anchors randomly placed within the environment. Bluetooth Low Energy (BLE) scans and WiFi scans were carried out at frequencies of 1 Hz and 3 Hz, respectively. 
We use Google's ARCore for ground truth which is an accurate Visual-Inertial SLAM software available on Android. 

As an overview of our method capabilities, we present in \cref{fig:multi-agent} the results of a 4-agents scenario walking through an environment using RoNIN as the pedestrian motion estimates and leveraging UWB ranging, as well BLE and WiFi coarse loop closing to correct the scale and drift errors.

\subsection{Adaptive Scaling using Range Measurements}

Our evaluation explores various parameters, including the number of UWB anchors, the initial scale value, and the initial position estimates of the UWB anchors. We perform evaluation with the four variants of our method: PDR, RoNIN, Adaptive PDR, and Adaptive RoNIN. The first two variants use a fixed scale value, while the latter two estimate the scale online. 
To exemplify the significance of adaptive scaling, we present a visual example of scale correction in \cref{fig:scaling-ex}, highlighting the system's ability to rectify an initial erroneous scale factor. 

First, we assessed the system's accuracy with respect to the number of UWB anchors employed. 
We tested the accuracy with all combinations of 0 to 4 anchors, 0 corresponding to the pedestrian motion estimates alone.
\cref{fig:nb-uwb} shows the mean and standard deviation results for the four variants of our method.
The results clearly demonstrate that accuracy increases with the number of anchors. Importantly, even a single anchor significantly enhances accuracy, showcasing the effectiveness of our system in environments with limited anchor availability. This is especially true for the adaptive versions of our pedestrian motion (i.e. Adaptive PDR and Adaptive RoNIN), as they can correct the scale of the trajectory even with a single anchor.
 Notably, RoNIN consistently outperforms PDR due to its reduced susceptibility to drift.
Next, in \cref{fig:scale_uwb}, we examined the influence of the initial scale value \(s\) (referenced in Eqs.~\ref{eq:pdr-translation}~and~\ref{eq:ronin-translation}) on accuracy. For fixed versions of pedestrian motion (PDR and RoNIN), an erroneous initial scale value can lead to a large error accumulation. Conversely, adaptive versions (Adaptive PDR and Adaptive RoNIN) estimate scale online and correct for poor initializations. 
Our evaluation also examined the impact of initial position estimates of UWB anchors. Beginning with accurate anchor positioning and introducing increasing levels of white Gaussian noise, Figure \ref{fig:uwb-placement-noise} demonstrates that our system's accuracy remains robust, as it estimates anchor positions as part of the optimization process. We compared our results with a standard range-only approach~\cite{zafariSurveyIndoorLocalization2019} which necessitates accurate initial anchor positioning to acheive reasonable estimates. This experiments showcases the versatility of our system, which can be used in environments where anchor positions are unknown.

\subsection{Drift Correction via Coarse Relocalization}

A further contribution is the integration of BLE and WiFi-based coarse loop closing into our framework. We conducted experiments to assess the impact of this loop closing technique on the accuracy and robustness of our system.

\begin{figure}[ht]
    \centering
    \includegraphics[trim={4.0cm 1.5cm 2.3cm 2.5cm},clip,width=\columnwidth]{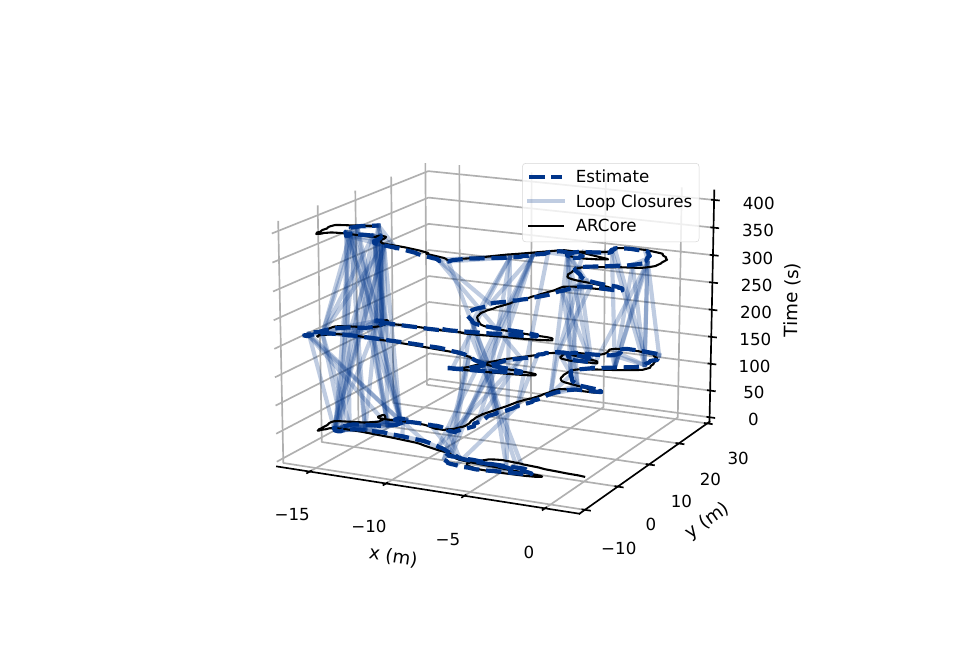}
    \caption{Illustration of the effect of BLE and WiFi-based coarse loop closing on localization accuracy. Our approach results in a RMSE of 1.05 $\pm$ 0.52, which is a significant improvement over RoNIN alone acheiving a RMSE = 2.88 $\pm$ 1.55.}
    \label{fig:ble-wifi-3d}
\end{figure}

As detailed in \cref{sec:coarse-proximity}, our loop closing method defines a circular region where loop closures incur no cost. Beyond this region, the cost increases as a function of distance. To account for the low scan rates of BLE and WiFi, we set the radius of this region to 2 meters. 
The experimental results depicted in \cref{fig:ble-wifi-3d} showcase the efficacy of this approach.
To evaluate the practical utility of BLE and WiFi loop closing, we conducted ten trajectories within an indoor environment, excluding UWB ranging. The resulting average accuracy, expressed as RMSE (m) and standard deviations, are summarized in Table \ref{tab:loops}. We compared against the standard proximity-based loop closing, which assumes that two steps poses linked by a loop closure are at the exact same position. Conversely, our coarse loop closing method that is better adapted to the low scan rate allowed on commercial phones, the step discretization, and the inherent inaccuracies of fingerprinting such as multi-path effects.

\begin{table}[t]
    \vspace{4mm}
    \centering
    \caption{RMSE (m) achieved with BLE and WiFi loop closing over 10 runs. Comparison between standard proximity loop closing and our approach. Smaller values are better.}
    \ra{1.2}
    \begin{tabular}{@{}llcccr@{}}\toprule
        \phantom{} & \phantom{}   & BLE & WiFi & BLE and WiFi& \phantom{}\\
        \cmidrule{3-5}
     \vspace{-3mm}
      \begin{filecontents*}{figures/ble_wifi.csv}
odom,loop,ble,wifi,blewifi
PDR,Proximity,2.39$\pm$0.96,3.31$\pm$1.61,3.22$\pm$1.33
,PEOPLEx,1.46$\pm$0.60,1.63$\pm$0.66,1.46$\pm$0.58
RoNIN,Proximity,2.38$\pm$1.06,2.68$\pm$1.22,2.85$\pm$1.22
,PEOPLEx,1.14$\pm$0.49,1.23$\pm$0.53,1.10$\pm$0.48
\end{filecontents*}
\csvreader[head to column names]{figures/ble_wifi.csv}{}{%
     \\\odom & \loop & \ble & \wifi & \blewifi&}%
    \\\bottomrule
    \end{tabular}
    \label{tab:loops}
    \vspace{-5mm}
\end{table}

The results demonstrate a substantial improvement in accuracy when employing our coarse loop closing method in conjunction with imperfect BLE and WiFi data. More importantly, as can be expected, naively performing proximity loop closing degrades the accuracy of the estimates. 
Therefore, while direct proximity-based approaches, such as the one used in~\cite{liuCollaborativeSLAMBased2020}, can achieve good performance when the measurement rate is at least as frequent as the user steps, they are vulnerable to large errors when the scanning rate is too low. This motivates the need for our more permissive coarse loop closing mechanism.

\section{Conclusion}

In conclusion, this paper introduces a novel framework for pedestrian localization that capitalizes on an opportunistic multi-sensor approach, leveraging IMU-based inertial navigation as the backbone and integrating UWB, BLE, and WiFi when available to enhance accuracy. 
The framework 
requires no prior environmental knowledge, and incorporates novel factors for adaptive scaling and coarse loop closures. Experimental validation using commercial smartphones in real indoor environments demonstrates the effectiveness of our approach. 
Future work will explore activity-aware localization, on-device estimation, and the integration of additional sensing signals, such as WiFi RTT or CSI, in our framework.

\newpage
\bibliographystyle{IEEEtran}

\vfill

\end{document}